\documentclass{ifacconf}%
\usepackage{graphicx}      
\usepackage{natbib}        
\graphicspath{{figures/}}  %
\usepackage{siunitx}      
\usepackage{multirow}      
\usepackage{subfigure}%
\usepackage{microtype}%
\usepackage{amsmath}       
\usepackage{booktabs,xcolor,siunitx}
\definecolor{lightgray}{rgb}{0.850980392156,0.85490196078, 0.85882352941176}

\usepackage{pgfplots}%
\usepackage{soul}%
\pgfplotsset{compat=newest}%
\usepgfplotslibrary{groupplots}%
\usepgfplotslibrary{dateplot}%

\newcommand\copyrighttext{%
  \footnotesize \copyright \quad 2020 the authors. This work has been accepted to IFAC for publication under a Creative Commons Licence CC-BY-NC-ND}
\newcommand\copyrightnotice{%
\begin{tikzpicture}[remember picture,overlay]
\node[anchor=north,yshift=0pt] at (current page.north) {\fbox{\parbox{\dimexpr\textwidth-\fboxsep-\fboxrule\relax}{\copyrighttext}}};
\end{tikzpicture}%
}
\begin{document}%
\begin{frontmatter}%
\title{Modeling System Dynamics with Physics-Informed Neural Networks Based on Lagrangian Mechanics\thanksref{footnoteinfo}}%
\thanks[footnoteinfo]{This work was sponsored by the German Federal Ministry of Education and Research (ID: 01\,IS\,18049\,A).}%
\author[First,Second]{Manuel A. Roehrl}
\author[First,Second]{Thomas A. Runkler}
\author[First]{Veronika Brandtstetter}
\author[First]{Michel Tokic}
\author[First]{Stefan Obermayer}
\address[First]{Siemens AG, Corporate Technology, 81739 Munich, Germany (e-mail: manuel.roehrl@siemens.com; thomas.runkler@siemens.com; veronika.brandstetter@siemens.com; michel.tokic@siemens.com)}%
\address[Second]{Technical University of Munich, 85748 Garching, Germany \\(e-mail: m.roehrl@tum.de; thomas.runkler@tum.de)}%
\begin{abstract} 
Identifying accurate dynamic models is required for the simulation and control of various technical systems. In many important real-world applications, however, the two main modeling approaches often fail to meet requirements: first principles methods suffer from high bias, whereas data-driven modeling tends to have high variance. Additionally, purely data-based models often require large amounts of data and are often difficult to interpret.\par%
In this paper, we present \textit{physics-informed neural ordinary differential equations} (PINODE), a hybrid model that combines the two modeling techniques to overcome the aforementioned problems. This new approach directly incorporates the equations of motion originating from the Lagrange mechanics into a deep neural network structure. Thus, we can integrate prior physics knowledge where it is available and use function approximation---e.\,g., neural networks---where it is not. The method is tested with a forward model of a real-world physical system with large uncertainties. The resulting model is accurate and data-efficient while ensuring physical plausibility.\par%
With this, we demonstrate a method that beneficially merges physical insight with real data. Our findings are of interest for model-based control and system identification of mechanical systems.
\end{abstract}%
\end{frontmatter}%
%
\section{Introduction}%
\copyrightnotice
Performant and expressive computer simulation models, able to seamlessly incorporate physical measurements, are key for \textit{digital twins}; these, in turn, are indispensable for the digitalization of industry \citep{rosen_about_2015}. This artificial replica refers to a virtual description of a system that can integrate models with data in real time. The combination enables what--if analyses in an artificial environment and optimization of processes and products without inferring the real process. Recently, there has been impressive progress in the field of data-based modeling \citep{NIPS2012_4824}, especially in the area of deep learning \citep{lecun_deep_2015}. Nevertheless, two factors are essential for this progress: (a) huge, high-quality data sets and (b) immense computing power resulting from clusters of arithmetic units. These resources have become available most recently (see Moore's law). Factor (a) allows variance reduction, and factor (b) enables larger models to be trained, which reduces bias \citep{fan_selective_2019}. However, one problem with this type of modeling is that there are not always large amounts of usable data available. If that is the case, data-based models often show high variance. Furthermore, complex models lose their interpretability because of the high number of adaptation parameters, whereas classical physical models usually only have a few parameters and thus remain understandable. In addition, physical models only require a small amount of data for calibration. Because of their simplicity, however, they usually have a high bias. To reduce model bias and bridge the gap between both model types,  various approaches have been employed, including learning of correction terms and semi-physical models based on different subsystems or multi-fidelity modeling \citep{fernandez-godino_review_2016, von_stosch_hybrid_2014}. One recent development is physics-informed neural networks \citep{raissi_physics-informed_2019-1, lutter_deep_2019, greydanus_hamiltonian_2019, zhong_symplectic_2019, gupta_general_2019, rackauckas2020universal}, which use mechanistic equations to endow neural networks with better prior. We follow this line and propose physics-informed neural ordinary differential equations (PINODE). Our approach uses the equations of motions to structure the neural network. The model is then integrated to obtain the final model output. Within this work, we investigate whether PINODE is applicable to a real system and is more accurate than a standard model derived from Lagrangian mechanics. Therefore, we demonstrate the procedure using a real-world mechanical benchmark system---an inverted pendulum mounted on a cart, or, in short, cart pole. We focus on only parametrizing non-conservative forces such as friction and using physical insights for the remaining parts of the differential equation.\par%
After this introduction, we describe the proposed methodology. We then explain the conducted experiment and discuss the resulting findings. The next section overviews related approaches to modeling system dynamics in a semi-physics manner. Lastly, we summarize the results and give an outlook on future work.%
\section{Methodology}%
In model-based control, dynamics of mechanical systems are modeled by linking the system state $\mathbf{q}$ with the acting input $\mathbf{u}$. Depending on the respective contexts either the forward $f$ or the inverse $f^{-1}$ model is used. We want to find a forward model%
\begin{equation}%
    f(\mathbf{q}, \dot{\mathbf{q}}, \mathbf{u})=\ddot{\mathbf{q}},%
    \label{eq:1}%
\end{equation}%
which simulates the system state change for a given input $\mathbf{u}$. To obtain a favorable coupling between input and states, we suggest first deriving the equations of motion with the \textit{Lagrange formalism} and then integrating them into a neural network structure.%
\subsection{Lagrangian Mechanics}%
Describing the trajectory of a system has been extensively studied, and various mathematical
formulations exist to derive the corresponding differential equations. Within this approach, a modified form of the Lagrange mechanics is used, i.\,e., the Euler--Lagrange formulation with generalized coordinates and non-conservative forces. The formalism uses the energy of the system; therefore, the Lagrangian $L$ is a function of the generalized coordinates $\mathbf{q}$, which is defined as%
\begin{equation}%
    L = T - V.%
    \label{eq:2}%
\end{equation}%
$T$ represents the entire kinetic energy and $V$ is the total potential of the system. Applying the calculus of variations yields the \textit{Euler--Lagrange equation} as follows:%
\begin{equation}%
    \frac{d}{d t} \frac{\partial L}{\partial \dot{\mathbf{q}}}-\frac{\partial L}{\partial \mathbf{q}}-\mathbf{Q}^{n c o n s}=0,%
    \label{eq:3}%
\end{equation}%
 where $\mathbf{Q}^{n c o n s}$ are the non-conservative forces. By inserting Equation~(\ref{eq:2}), the formula can be described as%
\begin{equation}%
    \frac{d}{d t} \frac{\partial T}{\partial \dot{\mathbf{q}}}-\frac{\partial T}{\partial \mathbf{q}}+\frac{\partial V}{\partial \mathbf{q}}-\mathbf{Q}^{n c o n s}=0,%
    \label{eq:4}%
\end{equation}%
since $V$ is not a function of $\dot{\mathbf{q}}$. Although Equation~(\ref{eq:4}) contains partial derivatives, the result of those derivatives provides \textit{ordinary differential equations} (ODEs) of the second order. By applying the chain rule, we can write the equations of motion in the common matrix form%
\begin{equation}%
    \mathbf{M}(\mathbf{q}) \mathbf{\ddot{q}}+\mathbf{C}(\mathbf{q}, \mathbf{\dot{q}})\mathbf{\dot{q}} + \mathbf{G}(\mathbf{q}) = \mathbf{Q}^{n c o n s}.%
    \label{eq:5}%
\end{equation}%
Here, $\mathbf{M}(\mathbf{q})$ represents the inertia matrix, $\mathbf{C}(\mathbf{q}, \mathbf{\dot{q}})$ is the Coriolis matrix, and $\mathbf{G}(\mathbf{q})$ are the conservative forces. By using this equation, any mechanical system with holonomic constraints can be described, e.\,g., coupled pendulums or robotic manipulators.%
\subsection{Incorporating Equations into Neural Net Structure}%
On the basis of the equation of motion~(\ref{eq:5}), a classical engineering approach would now begin with measuring or estimating the required parameters and forces. In contrast, a learning approach would abandon the equations and find a mapping between input and states directly from data. We want to combine both approaches, using the structure of the Euler--Lagrange equation and directly parametrizing parts of it.\par%
While enforcing this structure, we do not use a direct function to map from an initial state to the next one. Instead, we learn the underlying ODE. Therefore, to find a future state, we need to integrate the differential equation. That way, the model is memory and parameter efficient \citep{chen_neural_2018}. Furthermore, we can use position and velocity measurements for the parameter optimization and avoid the need to measure accelerations. To solve the ODE, we apply the \textit{Runge--Kutta} method of the fourth order \citep{runge_ueber_1895}, which is a standard method for fixed-time-step integration, although we assume the use of other solver schemes is also possible. Given the integration method and a system's initial generalized coordinates and velocities, we can solve the initial value problem. In this manner, we find the future system state after some step size $h$ at some time $t_{n+1} = t_n + h$. We reduce the problem and define $\mathbf{z}_n=[\mathbf{q}_n, \dot{\mathbf{q}_n}]^T$. The iterative scheme can then be described for the equations of motion as\par%
\begin{equation}%
    \mathbf{z}_{n+1} =\mathbf{z}_{n}+\frac{1}{6}\left(\mathbf{k}_{1}+2 \mathbf{k}_{2}+2 \mathbf{k}_{3}+\mathbf{k}_{4}\right),%
    \label{eq:6}%
\end{equation}%
with%
\begin{equation}%
\begin{aligned} %
    \mathbf{k}_{1} &= h \cdot g\left(t_n, \mathbf{z}_{n}, \mathbf{u}_{n}\right) \\ %
    \mathbf{k}_{2} &= h \cdot g\left(t_n + h/2, \mathbf{z}_{n}+\mathbf{k}_{1} / 2, \mathbf{u}_{n}\right) \\ %
    \mathbf{k}_{3} &= h \cdot g\left(t_n + h/2, \mathbf{z}_{n}+\mathbf{k}_{2} / 2, \mathbf{u}_{n}\right) \\ %
    \mathbf{k}_{4} &= h \cdot g\left(t_n + h, \mathbf{z}_{n}+\mathbf{k}_{3}, \mathbf{u}_{n}\right). \\ %
\end{aligned}%
\label{eq:7}%
\end{equation}%
$\mathbf{g}(t,\mathbf{z},\mathbf{u})$ is a vector of the generalized velocities and generalized accelerations. The latter can be found by solving Equation~(\ref{eq:5}) for $\ddot{\mathbf{q}}$ (see Equation~(\ref{eq:9})).\par%
We propose the use of a universal function approximator, like a neural network, for parts, that are unknown or nontrivial to model but define the physical parameters where knowledge is available. For example, parameters like gravity, lengths, masses and moments of inertia  are often easy to identify, whereas non-conservative forces are more difficult to determine. It is nontrivial because the forces often represent one or more partly interfering physical phenomena, e.\,g., friction, air drag, or fluid interactions. Models used to describe those phenomena are often not derived from first principles and need the use of extensive empirical methods, which are sometimes either specific for an individual application or only poorly approximate the underlying physical phenomena, or both.\par%
Generally, identifying physical parameters jointly with the neural network has only worked for a very limited parameter space. Therefore, if a certain part of the differential equation is not known, this part should be replaced by a universal approximator like e.\,g. a multi-layer perceptron.\par%
\begin{figure}[htb]%
    \centering%
    \begingroup%
        \def\svgwidth{0.42\textwidth}%
        \fontsize{8}{8}\selectfont%
        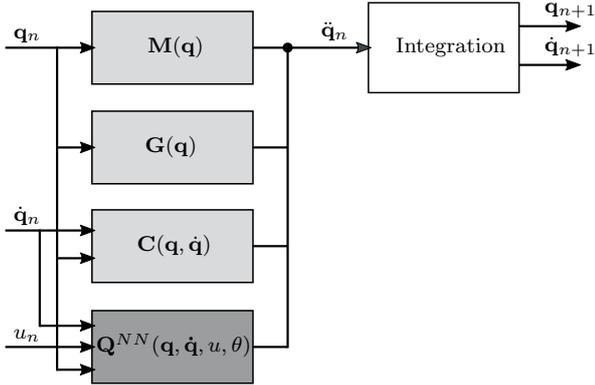%
    \endgroup%
    \caption{Simplified computation graph for one forward pass of an exemplary PINODE model to predict $\mathbf{q}_{n+1}$ and $\dot{\mathbf{q}}_{n+1}$ at time $t_{n+1}=t_{n} + h$.}%
    \label{fig:modelgraph}%
\end{figure}%
\newpage%
Figure \ref{fig:modelgraph} illustrates the technique described above for a single forward pass with a simplified computation graph. The diagram shows an example where the non-conservative forces are substituted by a neural network, which is described as %
\begin{equation}%
    \mathbf{Q}^{n c o n s} =\mathbf{Q}^{NN}(\mathbf{q}, \mathbf{\dot{q}}, \mathbf{u}; \theta),%
    \label{eq:8}%
\end{equation}%
where $\theta$ represents the trainable weights. The remaining matrices $\mathbf{M}$, $\mathbf{C}$, and $\mathbf{G}$ are determined by traditionally measuring or estimating the necessary parameters. All components together are solved to yield the explicit form for the acceleration:%
\begin{equation}%
    \mathbf{\ddot{q}} = \mathbf{M}(\mathbf{q})^{-1} \Bigl[\mathbf{Q}^{NN}(\mathbf{q}, \mathbf{\dot{q}}, u, \theta)- \mathbf{C}(\mathbf{q}, \mathbf{\dot{q}}) \mathbf{\dot{q}} - \mathbf{G}(\mathbf{q})\Bigr].%
    \label{eq:9}%
\end{equation}%
We then apply the integration method, described in Equation~(\ref{eq:6}), to obtain the position $\mathbf{q}_{n+1}$ and velocity $\mathbf{\dot{q}}_{n+1}$ for the next state.%
\subsection{Learning Parameters from Time Series Data}%
Having discussed how to construct the PINODE model, we next address how to obtain the network variables $\theta$. For this, the optimization problem is defined as follows:%
\begin{equation}%
    \theta^* = \min \mathcal{L}\bigl(\text{ODESolve}\bigl( f(\mathbf{q}, \dot{\mathbf{q}}, \mathbf{u}; \theta)\bigr), \mathbf{q}, \mathbf{\dot{q}}\bigr),%
    \label{eq:10}%
\end{equation}%
where $\mathcal{L}$ can be an arbitrary loss function. The only constraint is that it must be differentiable with respect to its parameters to enable gradient computation with \textit{automatic differentiation}. In particular, we use the reverse-mode automatic differentiation to get the derivatives of the loss in Equation~(\ref{eq:11}) toward its weights $\theta$. However, for the experiment in the subsequent section, we used the following cost function:%
\begin{equation}%
    \mathcal{L}(\theta) = \frac{1}{N} \sum_{k=1}^{N}\mathbf{\lambda}_1( \mathbf{q}_{k} - \mathbf{q}_{k}^{\prime} )^{2} + \mathbf{\lambda}_2 ( \dot{\mathbf{q}}_{k} - \dot{\mathbf{q}}_{k}^{\prime} )^{2},%
    \label{eq:11}
\end{equation}%
where the $k$th future state is found by integrating Equation~(\ref{eq:9}):%
\begin{equation}%
   [\mathbf{q}_k', \dot{\mathbf{q}}_k'] = \text{ODESolve}\bigl( f(\mathbf{q}_k, \dot{\mathbf{q}_k}, \mathbf{u}_k; \theta)\bigr),%
\end{equation}%
$N$ represents the number of state transitions for which the loss is calculated. The $\lambda$-factors can be chosen to put weight on specific generalized coordinates or velocities. If a generalized degree of freedom is described with angular coordinates, the loss function must be slightly extended. This is because the fact that angles lie in a non-Euclidean space, mostly between $[0, 2\pi]$, which impedes the optimization. To address this issue, we use, instead of the direct angle  $\varphi$, an embedded form $(cos(\varphi) + sin(\varphi))$ for the cost calculation. Finally, we find the parameters with the Adam optimizer with a learning rate of $10^{-3}$ \citep{kingma_adam_2014}, although the use of other gradient descent algorithms is generally also possible.%
\section{Experiments}%
\begin{figure}%
    \centering%
    \subfigure[Schematic drawing of the test rig.]{%
    \begingroup%
        \def\svgwidth{0.28\textwidth}%
        \fontsize{8}{8}\selectfont%
        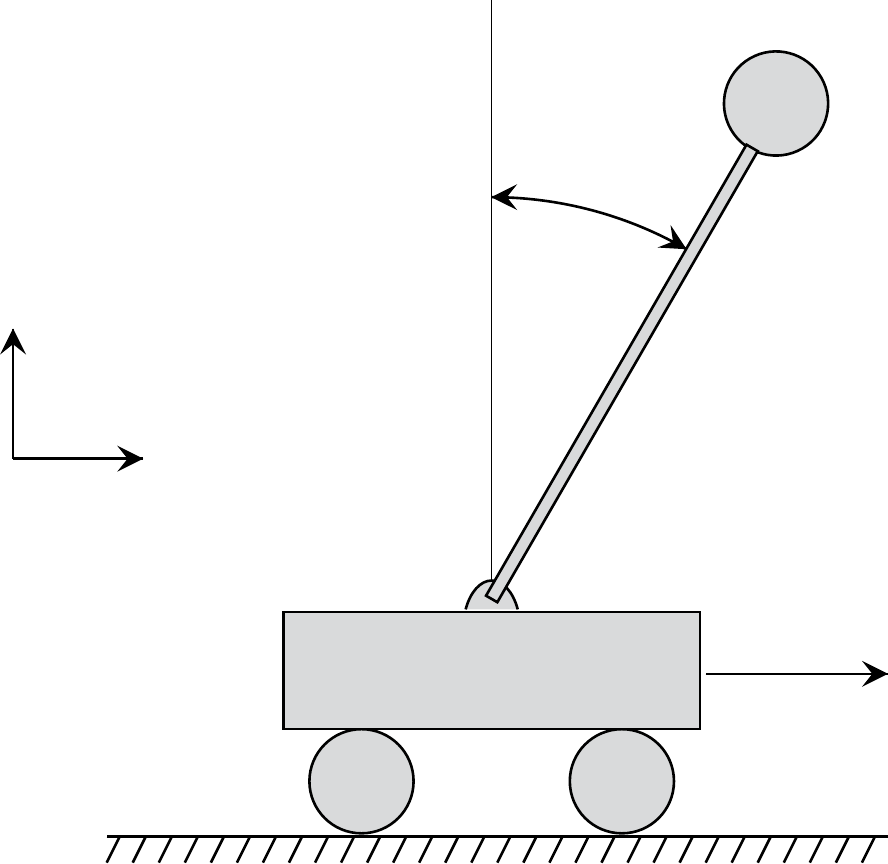%
    \endgroup}
    \qquad%
        \subfigure[Picture of the test rig system, presented at CeBIT 2018.]{%
    \includegraphics[width=0.45\textwidth]{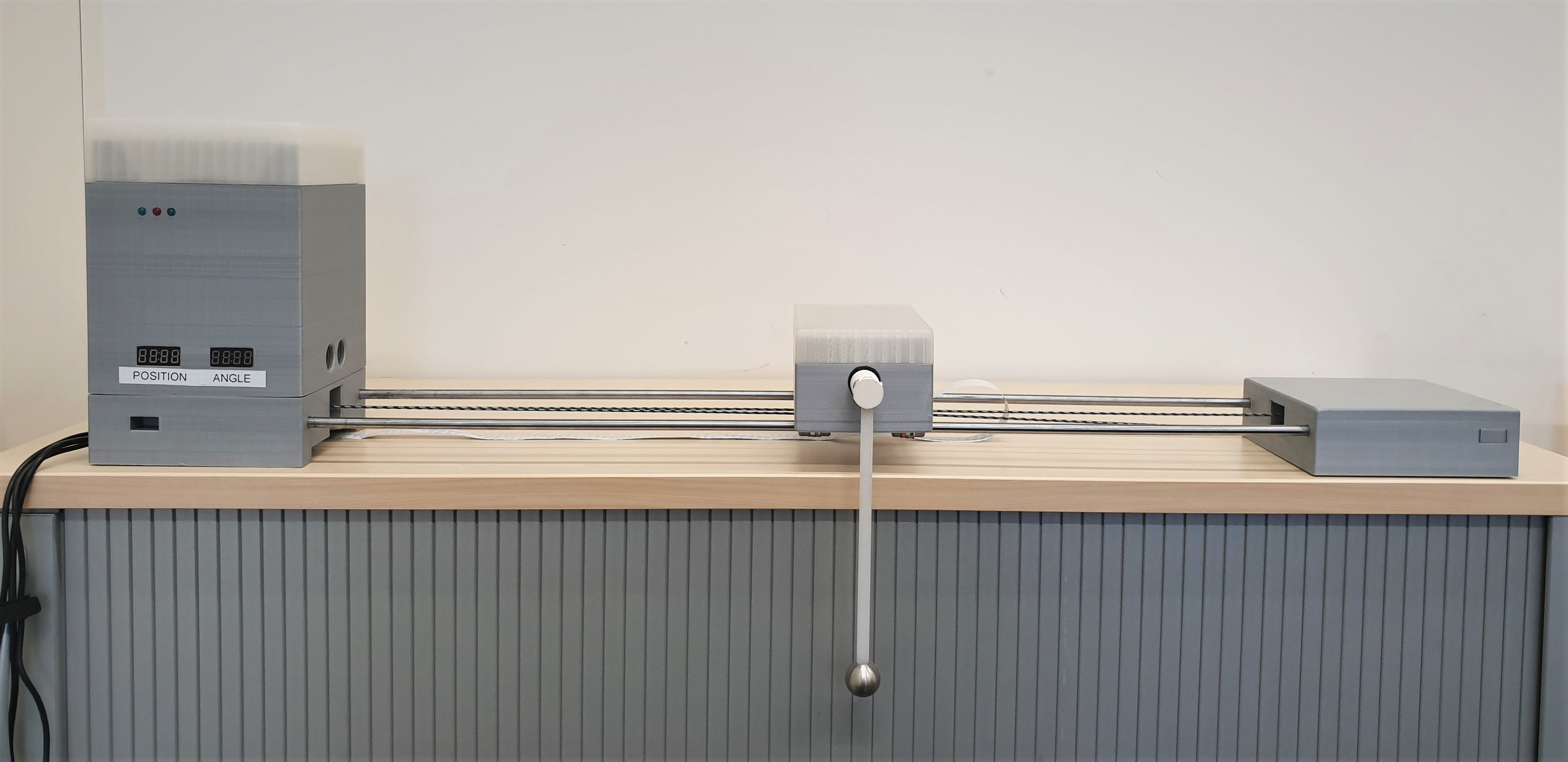}}%
    \caption{Schematic drawing and picture of the test rig system---the inverted pendulum on a cart.}%
    \label{fig:cartpole}%
\end{figure}

For the purpose of testing and evaluation, we used the method for system identification of a real-world physical cart pole. The subject of investigation is shown by a schematic sketch and a photograph in Figure \ref{fig:cartpole}. %
We want to justify the claim that the approach can learn the non-conservative forces of a real complex system. To proof that, we show that it is possible to integrate existing knowledge seamlessly and to map the physical, still un-modeled components with the neural network. \par%
\subsection{Experimental Setup}%
The cart is connected with a string to a motor, which can be actuated to give control impulses to the system. The control input moves the cart along the linear guidance, which has a total length of about \SI{0.4}{\meter}. The test rig is equipped with two sensors that record the position of the cart and pole. The first is measured with an ultrasonic distance sensor, and the second by an optical hollow shaft encoder. Velocities are calculated from the difference between two measurements. Additionally, the introduced control input is logged and saved with the state measurements. Within the experiment, trajectories were generated by manually controlling the cart in a random fashion. In total, samples of \SI{8}{\minute} measuring were used for training, this corresponds to approximately 200 times moving from right to left and back again. The data are recorded at a sample frequency of \SI{50}{\hertz}. Accordingly, the time step of the integrator was set to the resulting time interval of \SI{0.02}{\second}. Table \ref{tab:data} gives an overview of information about the data used for training. \par%
The algorithm is implemented within the \textit{TensorFlow} framework, which enables automatic differentiation and offers various optimizers and predefined modules for building deep learning architectures. The models created this way use the same parameters, except for the friction factor of the cart, which is needed only for the pure ODE comparison model. All used parameters of both models are specified in Table \ref{tab:para}.\par%
\begin{table}[htb]%
    \begin{center}%
    \caption{Information about the training data.}%
    \label{tab:data}%
    \begingroup\setlength{\fboxsep}{0pt}%
    \colorbox{lightgray}{%
    \begin{tabular}{llrrrr}%
    \hline%
     Sample rate &\multicolumn{5}{c}{\SI{50}{\hertz}} \\%
     Samples & \multicolumn{5}{c}{24,271} \\  \hline%
     \multirow{6}{0.07\textwidth}{Statistics}& & & & &\\
      &                        & Mean  & STD.  & Min    & Max \\
      &$x$ (m)                 & $0.024$ & $0.126$ & $-0.325$ & $0.276$\\
      &$\varphi$ (rad)         & $3.290$ & $1.823$ &  $0$     & $6.260$\\
      &$\dot{x}$ (m/s)         & $0$     & $0.430$ & $-4.398$ & $4.141$\\
      &$\dot{\varphi}$ (rad/s) & $-0.328$ & $6.742$ & $-23.562$ & $18.326$\\ \hline%
    \end{tabular}%
    }\endgroup%
    \end{center}%
\end{table}%
\begin{table}[htb]%
    \caption{Parameters describing dynamics of cart and pendulum apparatus}%
    \begin{center}%
    \label{tab:para}%
    \begingroup\setlength{\fboxsep}{0pt}%
    \colorbox{lightgray}{%
    \begin{tabular}{lrl}%
    \hline%
    Parameter & Value  & Unit\\\hline%
    mass cart $m_c$ & $0.466$ & \si{kg}\\%
    mass pole $m_p$    & $0.06$ & \si{kg} \\ 
    mass sphere $m_s$ & $0.012$ & \si{kg} \\%
    length pole $l$ & $0.201$ & \si{m} \\%
    friction factor cart $\mu_c$ & $0.0408$ & - \\%
    friction factor pole $\mu_p$ & $0.0020$ & - \\
    gravity $g$ & $9.81$ & \si{m/s^2} \\ \hline%
    \end{tabular}%
    }\endgroup%
    \end{center}%
\end{table}%
\subsection{PINODE Model}%
The parameters gravity, lengths, masses, and moments of inertia are measured for the cart pole. With this, the matrices $\mathbf{M}$, $\mathbf{C}$, and $\mathbf{G}$ are defined. Further, we use viscous friction for the pole bearing. The residual non-conservative forces $\mathbf{Q}$ are modeled by a artificial neural network depending on the last system state and control input $u$. The model is composed of a multi-layer perceptron with two hidden layers each containing 50\,units. All layers beside the last one use rectifier activation functions; the last applies a hyperbolic tangent function. The model has five inputs and one output, which gives a total of 5,451\,trainable parameters. Before training, the samples were shuffled randomly and then combined into batches of 128 elements. \par%
\subsection{Pure ODE Model}%
The performance of the PINODE model is compared with that of a standard model derived with the Lagrange formalism (\ref{eq:4}). This model contains non-conservative forces to consider both friction between cart and linear guidance, and resistance in the pole bearing. For the former, Coulomb's friction with constant normal force, and for the latter, viscous friction was used. Thus, the block for the non-conservative forces in Figure \ref{fig:modelgraph} is described for the pure ODE model as follows:\par%
\begin{equation*}%
    \mathbf{Q}^{n c o n s} =%
\end{equation*}%
\begin{equation}%
   = \left[\begin{array}{c}    {u-(m_P + m_K + m_C) \cdot g \cdot \mu_C \cdot sign(\dot{x})}  \\ %
    {-\mu_P \cdot \dot{\varphi}} \end{array}\right].%
\label{eq:13}%
\end{equation}%
The necessary friction factors were found by least squares optimization.\par%
\section{Results and Discussion}%
\begin{figure}[htb]%
    \centering%
    %
    \begingroup%
    \fontsize{6}{6}\selectfont%
        \input{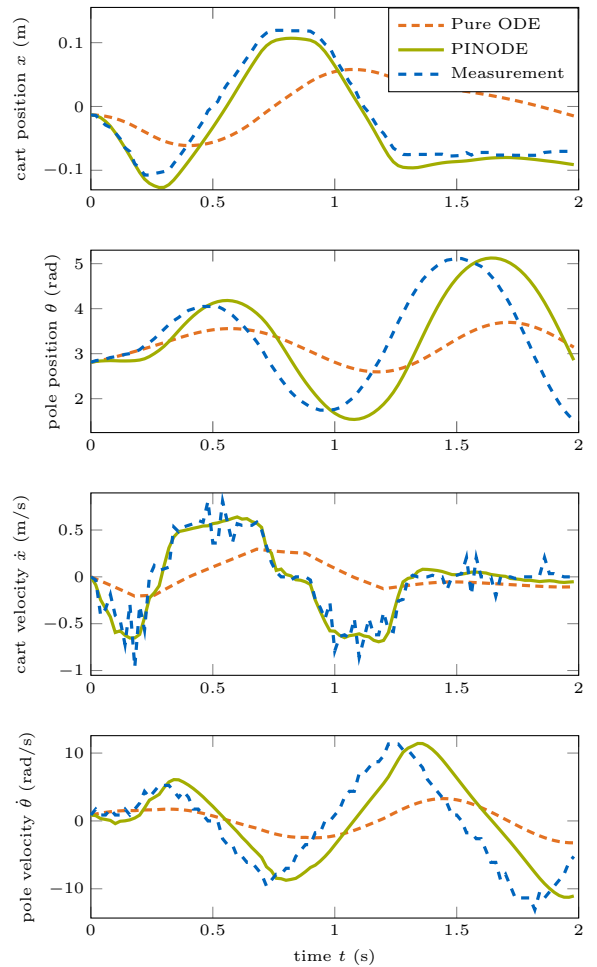}%
    \endgroup%
    \caption{Pure ODE simulation and PINODE model predictions for \SI{2}{\second} with the same initial conditions and control input $u$ compared with sensor data.}%
    \label{fig:trajectoy}%
\end{figure}

Figure \ref{fig:trajectoy} compares the measured ground truth to the pure ODE model and the PINODE model by showing simulations for the generalized coordinates over hundred time steps (2\,seconds), while starting from the same initial state. The cart position of the hybrid model in the top left corner of the figure is predicted very accurately---only a small error occurs. Similarly, the cart velocity is reproduced precisely, although the measurements are very noisy. On the right side of the figure, the pole positions $\varphi$ and velocities $\dot{\varphi}$ of the PINODE model show that a small error is accumulating over time, but not to the same extent as in the case of the cart position $x$ and velocity $\dot{x}$ on the left. The pure ODE curves in all four graphs show a far greater deviation from the measured trajectory than does the semi-physical model, although the principle tendency is correct.\par%
\begin{figure}[htb]%
    \centering%
    %
    \begingroup%
        \resizebox{0.48\textwidth}{!}{\input{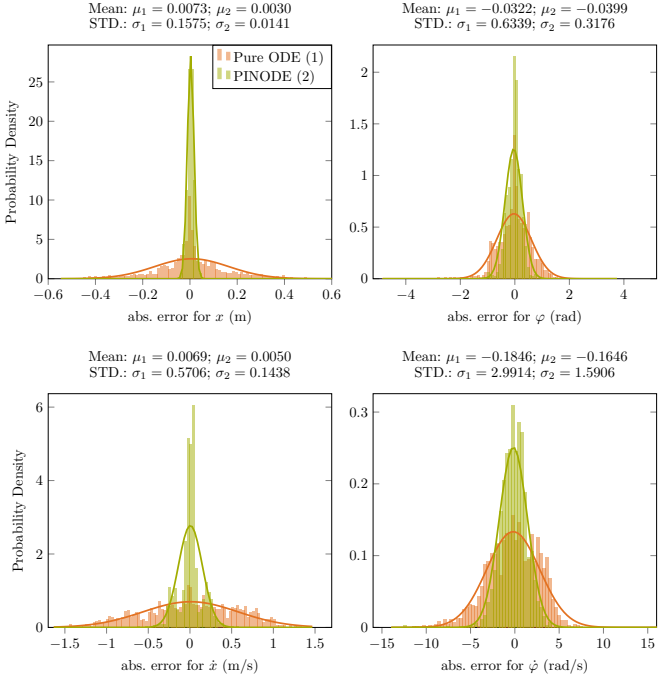}}%
    \endgroup%
    \caption{Histograms showing absolute errors of pure ODE and PINODE model generalized coordinates and velocities for one thousand 0.6\,seconds-long simulations. A normal distribution is fitted and plotted for each histogram, their means $\mu_{\{1,2\}}$ and standard deviations $\sigma_{\{1,2\}}$ are shown above the respective graphs.}%
    \label{fig:hist}%
\end{figure}%
To present the performance of the approach for a larger state space, Figure \ref{fig:hist} illustrates the absolute errors over multiple simulated time intervals. More specifically, the models were evaluated for thirty time steps for thousand times. The initial state was set to the real measurement in every time interval. For each time frame and state, the absolute error between both models and the measured ground truth was calculated. The resulting error values are plotted as histograms. This representation confirms the observation from the time sequences in Figure \ref{fig:trajectoy}: the kinematics of the cart are modeled more accurately than are those of the pole. A possible explanation for this might be that the pendulum has more complex dynamics than has the cart. Overall, the PINODE model completely outperforms the pure ODE model. The better performance of the composite model must be attributed to the universal approximation capability of the neural network component, because it is able to take phenomena into account, which have not been considered in the pure ODE model. A few examples that could be observed as phenomena in the real-world system are friction depending on the $x$-direction, elasticity in the string connecting cart and motor, elasticity in the linear guidance, time delay of the control input and mechanical clearance in the bearings.\par%
In addition to the two described models, a standard---black box---neural network was trained to map from an initial state to the hidden state. The performance of this model is not shown in Figures \ref{fig:trajectoy} and \ref{fig:hist} because it was unable to learn the dynamics with the same training data. We presume this is because the proposed structure contains fewer free parameters and therefore requires less data than a pure black box model. In addition, the PINODE model ensures more physical interpretability.\par%
\section{Related Work}%
Research on modeling of dynamics has a long history. Models can be learned from data, derived from physics, or developed in a hybrid, i.\,e., semi-physical, manner \citep{ljung_modeling_1994}. When deriving the model from physics, the dynamics parameters are either estimated or calibrated, often by using least-squares regression. The data-driven models mostly use standard representation learning methods to fit a model in the proximity of the available data. However, this paper follows the line of hybrid modeling, namely, the combination of  differential equations with neural networks.\par%
\textbf{Learning Differential Equations\quad} Methods relying on the universal function approximation ability of machine learning methods to solve ordinary or partial differential equations (PDEs) have already been discovered early \citep{lagaris_artificial_1998}, but have been rediscovered recently \citep{raissi_physics-informed_2019-1, long_pde-net_2017}. These studies focus on using feed-forward networks to overcome limitations of differential equation solvers by designing the loss function according to the respective equations and taking advantage of the efficient derivative computation in neural networks. Rather than solving the ODEs, our work focuses on structuring the network and modeling the underlying change rate of the physical process.\par%
\textbf{Differentiating through ODE Solver\quad} Much recent work has proposed integrating an ODE solver into the network structure. \cite{chen_neural_2018} propose a general method to parameterize the derivative of the hidden state and then apply an arbitrary ODE solver. \cite{gupta_general_2019}, \cite{greydanus_hamiltonian_2019} and \cite{zhong_symplectic_2019} take such a perspective for mechanical systems and model the derivative of the desired state. An earlier example for this idea, applied in another domain, given by \cite{al_seyab_nonlinear_2008}. Their approach parameterizes dynamic sensitivity equations with a recurrent neural network and then integrates the ODE with Taylor series.\par%
\textbf{Structuring Learning Problems with Physical Prior\quad} A number of studies have suggested endowing neural networks with better physical prior. One concurrent work by \cite{rackauckas2020universal} suggests a general semi-mechanistic approach where part of the differential equation is an universal approximator. Their approach shares a similar motivation to ours and also achieves improved data and computational efficiency in diverse examples. Two recent works aim to uncover physical laws from data in a general manner \citep{iten_discovering_2018, greydanus_hamiltonian_2019}. More specific for mechanical systems, physics-informed neural networks were demonstrated with simulated time series data for the forward model (\ref{eq:1}) of a pendulum, double pendulum, and a cart pole system \citep{gupta_general_2019, zhong_symplectic_2019}. Instead of using generalized coordinates, \cite{zhong_symplectic_2019} use translational coordinates to avoid the problem, that comes with angle data. An example for the application to a simulated and real robot system is given by \cite{lutter_deep_2019}. They use a similar approach, but learned the inverse model. Further they also learned the forward model for a physical Furuta pendulum \citep{lutter_deep_2019-1}. In contrast to our setup, they had measurements for second-order derivatives, which is why it was not necessary to differentiate through an ODE solver. \par%
\section{Conclusion}%
The most obvious finding to emerge from this study is that, for the forward model of a real-world physical system, it is possible to integrate existing parts of the equation of motion and model residual physical effects, which are not trivial to capture, by a neural network. Compared with a complete black box approach, the method needs less data, and a certain physical interpretability is retained. We can answer the research question conclusively and demonstrate with PINODE the usage of equations of motion as model prior. Therefore, we can propose a further step toward bridging the gap between purely data-driven and mechanistic models. We have thus taken up the related work by applying and adapting recent methods to develop a forward model for a physical cart pendulum system. \par%
In future works, the technique may be compared with more advanced system identification approaches and studied for its ability to extrapolate. Furthermore one could consider to use different solvers and automatically switch between implicit and explicit methods depending on whether the problem is stiff or non-stiff. Besides that, to further investigate the generality of the approach, additional experiments should be performed and the method adapted for the application to more complex systems, where ODEs or PDEs exist. Other possible extensions to this work may be to use the forward model as environment for learning reinforcement policies \citep{hein2018interpretable}. \par%
\begin{ack}%
The contribution was supported with funds from the German Federal Ministry of Education and Research within the project ``ALICE-III: Autonomous Learning in Complex Environments'' under the identification number 01\,IS\,18049\,A.%
\end{ack}%
\bibliography{ifacconf}             
%
\appendix%
\end{document}